\documentclass[10pt,twocolumn,letterpaper]{article}

\usepackage[pagenumbers]{cvpr} %

\usepackage{multirow}
\usepackage{makecell}
\usepackage{xspace}
\usepackage{adjustbox}

\newcommand{\ours}{TinyAstral\xspace}
\newcommand{\oursfour}{TinyAstral-4B\xspace}

\definecolor{cvprblue}{rgb}{0.21,0.49,0.74}
\usepackage[pagebackref,breaklinks,colorlinks,allcolors=cvprblue]{hyperref}

\def\confName{CVPR}
\def\confYear{2027}

\title{TinyAstral: Transferring Self-Evolving Harnesses\\to Tiny Generalist Manipulation Agents}
\hypersetup{pdftitle={TinyAstral: Transferring Self-Evolving Harnesses to Tiny Generalist Manipulation Agents}}

\author{%
Kehong Gong$^{1}$\quad
Xiaoxu Zheng$^{1}$\quad
Guanli Hou$^{2}$\quad
Seunghun Lee$^{1}$\quad
Mengqi Guo$^{1}$\\
Runjia Tan$^{3}$\quad
Zhengyu Wen$^{2}$\quad
Qi Wang$^{2}$\quad
Xiaoyu He$^{2}$\quad
Mingyuan Zhang$^{2,*}$\quad
Michael Bi Mi$^{1}$\\[0.6ex]
{\normalsize $^{1}$Huawei Technologies Co., Ltd\qquad $^{2}$Huawei Noah Research Institute\qquad $^{3}$Independent Researcher}\\
{\small $^{*}$Corresponding author}
}

\begin{document}
\twocolumn[{%
\renewcommand\twocolumn[1][]{#1}%
\maketitle
\begin{center}
\includegraphics[width=\textwidth]{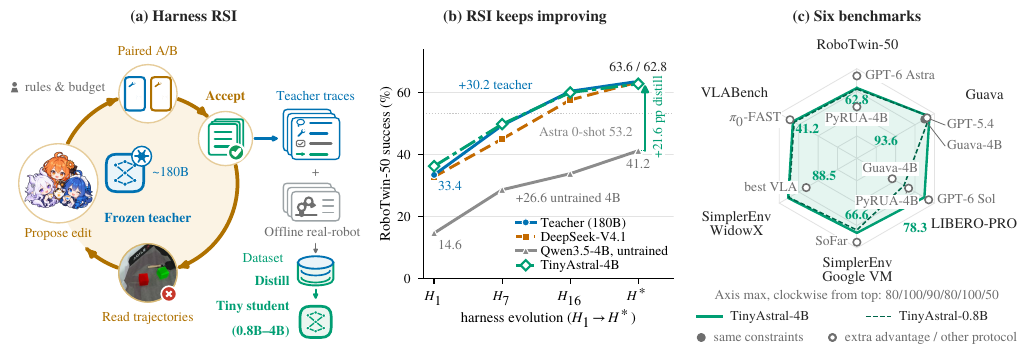}
\vspace{-1.5ex}
\captionof{figure}{\textbf{\ours transfers harness evolution gains to tiny robot agents.}
(a)~With teacher weights frozen, RSI improves the harness; we then freeze the final harness, collect teacher trajectories, and distill small students.
(b)~Harness evolution improves the teacher and other frozen planners. Students trained separately at four harness versions each approach their corresponding teacher.
(c)~Final students on training and unseen benchmarks. Solid/dashed lines: 4B/0.8B students; filled/hollow points: comparisons under the same constraints versus additional advantages or different protocols (\cref{tab:main}). Axes are scaled independently.}
\label{fig:teaser}
\end{center}
\vspace{1ex}

}]

\begin{abstract}
Robot agents built on vision--language models depend on both the planner and its harness: the tools, feedback, memory, and recovery logic around it. Can improvements to a large teacher's harness be carried into a compact agent with lower inference compute cost?
We present TinyAstral, a two-stage approach. We first freeze an open-weight teacher and improve its harness through recursive self-improvement (RSI): frontier LLM agents read the teacher's trajectories and propose harness edits, and each edit is accepted only if it passes a paired statistical test. We then freeze the final harness, collect teacher trajectories with reasoning and failure recovery, and distill a small student that replaces the teacher in the same harness.
On RoboTwin, harness evolution raises the teacher's success from 33.4\% to 63.6\%; the final 4B student reaches 62.8\%.
Across four harness versions, distilled 4B students score close to their corresponding teachers, and later versions yield stronger students, showing that harness evolution gains carry into compact models. The harness gains also transfer to other frozen planners without retraining.
On identical episodes, the final 4B student outperforms open 4B agents, the open 27B harness-RSI model SenseNova-RoboRSI, and our own teacher model placed in two competitors' harnesses.
It stays close to its teacher on all three training benchmarks, and on three benchmarks never used for RSI or training it scores close to a 552B planner running the same harness. On the measured deployment episodes, it uses about one fifth of its teacher's accelerator time.
Project page: \href{https://garfield-kh.github.io/TinyAstral/}{\texttt{garfield-kh.github.io/TinyAstral}}
\end{abstract}

\section{Introduction}
\label{sec:intro}

Vision--language models (VLMs) can operate robot arms by reading images and calling perception and motion tools~\cite{liang2023code,huang2023voxposer,liu2024moka,fu2026capx}.
The model chooses the object, action sequence, and stopping point; the surrounding \emph{harness} grounds these choices, executes them, and reports what happened. Tool interfaces, feedback, task memory, and failure recovery therefore contribute to the agent's performance alongside the model itself~\cite{zhang2026harness,tian2026physevo,wang2026harnesspai,deng2026sensenova}.

Two lines of work offer complementary ways to build these agents. Harness evolution improves the system around a frozen planner~\cite{tian2026physevo,wang2026harnesspai,deng2026sensenova}; agent distillation trains compact models from a stronger agent's trajectories~\cite{liu2026guava,li2026robo,zhang2026world}.
We connect these approaches by asking: \emph{can we first improve a large teacher's harness, then transfer the resulting decision-making capability to a small model?}
This would allow the improved system to run with substantially less inference compute.

We present \textbf{TinyAstral}, a family of 0.8B--4B manipulation agents built in two stages (\cref{fig:overview}).
First, we freeze the weights of an open-weight teacher, Qwen3.8-Flash-Next~\cite{qwen2026qwen38}, and improve its harness through a master-guided recursive self-improvement (RSI) loop. Humans specify the rules and budget in advance. Frontier LLM agents then inspect failures and propose harness edits, which the system automatically accepts or rejects according to pre-registered paired statistical tests.
Second, we freeze the final harness and collect teacher trajectories containing reasoning, tool calls, and recovery from injected failures. Together with auxiliary offline real-robot supervision, these trajectories train a compact student to interpret feedback, choose actions, recover from failure, and decide when to finish. At deployment, the student replaces the teacher in the same frozen harness.

We evaluate both stages and the connection between them. Harness evolution raises the frozen teacher's RoboTwin success from 33.4\% to 63.6\%, and distillation produces a 4B student at 62.8\%.
To test whether this connection holds beyond the final endpoint, we distill students at four harness versions. The four students score 36.2\%, 49.8\%, 60.0\%, and 62.8\%, each close to its corresponding teacher.
The same evolved harness also improves DeepSeek-V4.1-Flash and the untrained 4B initialization, showing that its gains are reusable across planners. On identical episodes the student outperforms every open competitor we ran, including our own teacher model inside two competitors' harnesses (93.6\% vs.\ 78.9\% on Guava; 78.3\% vs.\ 57.3\% on LIBERO-PRO), and on unseen benchmarks the small 4B student scores close to a 552B planner in the same harness.

Our contributions are:
\begin{itemize}
\item \textbf{Harness evolution followed by distillation.} A two-stage approach first improves a frozen teacher's manipulation harness, then trains a compact planner from that teacher's reasoning and recovery trajectories for deployment in the same harness.
\item \textbf{Evidence that evolution gains transfer.} The harness improves other frozen planners, and distilled students at four harness versions track the performance of their corresponding teachers, showing that system improvements carry into compact models.
\item \textbf{Compact agents with lower inference compute cost.} A single TinyAstral-4B model covers six benchmarks: on identical episodes it outperforms every open competitor we ran, including our own teacher model inside two competitors' harnesses, scores close to its teacher on the three training benchmarks and close to a 552B planner on unseen ones, and uses about one fifth of the teacher's accelerator time per episode in our deployment measurement.
\end{itemize}
\section{Related Work}
\label{sec:related}

\paragraph{Large models operating robots through a harness.}
Language and vision-language models can operate robots by writing code or issuing tool calls over perception and motion primitives~\cite{liang2023code,huang2023voxposer,liu2024moka,fu2026capx}, with spatial-reasoning models supplying grounding~\cite{qi2025sofar,yuan2025embodiedr1}.
Recent work shows that the harness around the model, namely its tool interface, feedback, memory, and recovery logic, is a first-order factor~\cite{huang2026incontext}.
Harness VLA~\cite{zhang2026harness} composes a frozen VLA with memory-guided agents, RoboDawn~\cite{guo2026transferring} and MotorMind~\cite{li2026motormind} expose closed-loop interfaces to a general VLM, and PyRUA-Lean~\cite{si2026fewer} cuts token cost with interactive code execution.
A growing line also \emph{evolves} the harness from execution traces: PhysEvo~\cite{tian2026physevo} runs recursive self-improvement around a single frozen model, HarnessPAI~\cite{wang2026harnesspai} evolves an executable program, ASPIRE~\cite{lu2026aspire} accumulates a skill library, and the two RoboRSI systems~\cite{deng2026sensenova,wen2026roborsi} refine skills and agent components with validation before release. Concurrent with our work, SenseNova-RoboRSI~\cite{deng2026sensenova} also trains a 27B VLM on validated interaction experience, and World Action Agent~\cite{zhang2026world} distills the tool-call traces of its skill-evolved harness into a 9B VLM.
We share the view that the harness is what improves while model weights stay fixed.
TinyAstral differs in three ways: the harness of a locally hosted open-weight teacher evolves continually under paired statistical selection, its gains are measured across planners at successive snapshots, and the evolved teacher is distilled into 0.8B, 2B, and 4B students that stay close to it.

\paragraph{Small manipulation agents and VLAs.}
End-to-end VLAs such as OpenVLA~\cite{kim2024openvla}, $\pi_0$~\cite{black2024pi0}, and $\pi_{0.5}$~\cite{intelligence2025pi05} map observations directly to actions and are trained on large robot datasets, which makes them compact at inference but dependent on in-distribution demonstrations; benchmarks such as LIBERO-PRO~\cite{zhou2025liberopro} and RoboTwin~2.0~\cite{chen2025robotwin} probe how far they carry under perturbation.
A second route trains a small model to be the \emph{agent} instead.
Guava~\cite{liu2026guava} designs a manipulation harness and distills reasoning and recovery trajectories from a frontier VLM into a 4B open model, PyRUA-Lean~\cite{si2026fewer} trains a 4B coder with GRPO on 45 LIBERO-PRO instances, and Robo-Harness K1~\cite{li2026robo} and World Action Agent~\cite{zhang2026world} fine-tune small VLM students on harness traces.
TinyAstral first evolves the teacher harness and then freezes it for data collection and distillation. Distillation at several frozen snapshots tests whether the preceding harness improvements carry into student performance.

\paragraph{Self-improving agents and agent distillation.}
Language agents can improve through stored experience and reflection~\cite{wang2023voyager,shinn2023reflexion}, and recent systems let a model rewrite its own scaffold or code: STOP~\cite{zelikman2023self}, ADAS~\cite{hu2024automated}, G\"odel Agent~\cite{yin2024godel}, and the Darwin G\"odel Machine~\cite{zhang2025darwin}, following the classical notion of recursive self-improvement~\cite{good1966speculations}; GEPA~\cite{agrawal2025gepa} reflects on execution traces to evolve prompts.
ADAS also reports transfer of discovered agents across models and domains. TinyAstral evolves the tools, execution feedback, task memory, and recovery logic of a manipulation harness under paired statistical selection, and carries the evolved system into a tiny student.
Knowledge distillation~\cite{hinton2015distilling,sanh2019distilbert} and its agent variants, FireAct~\cite{chen2023fireact} and AgentTuning~\cite{zeng2023agenttuning}, transfer trajectories from a strong language agent to a smaller one on text and tool-use tasks; Distilling Step-by-Step~\cite{hsieh2023distilling} uses teacher rationales as extra supervision, and BetterTogether~\cite{soylu2024bettertogether} alternates prompt optimization with fine-tuning of the same model.
We supervise rationales together with tool calls inside a closed-loop visual interaction history, and distill from a teacher whose harness has already been evolved, into a student of a different size.
\section{Method}
\label{sec:method}
\begin{figure*}[t]
\centering
\includegraphics[width=\textwidth]{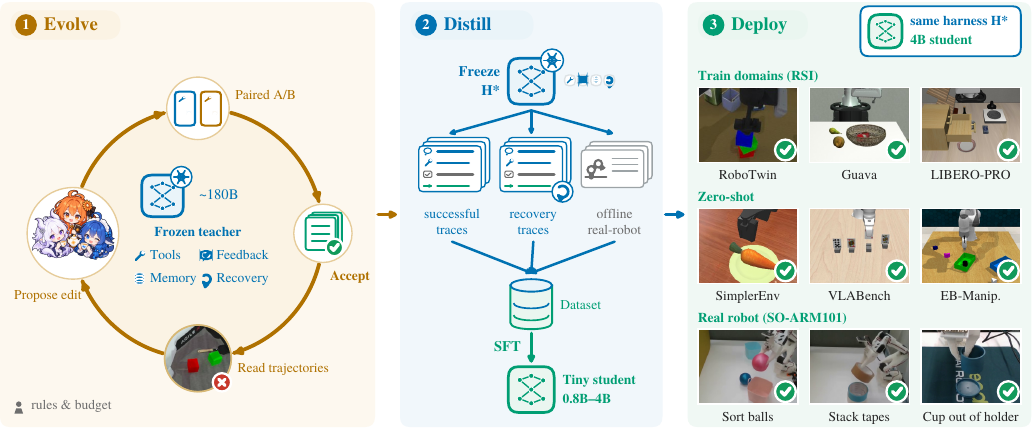}
\caption{\textbf{Overview of \ours.} Stage~1: humans specify the rules and budget in advance. A master of frontier LLM agents reads trajectories of the frozen teacher acting through the harness (tools, feedback, memory, recovery) and proposes harness edits, which are accepted only if a paired A/B test on selection seeds passes a pre-registered statistical gate; model weights never change. Stage~2: after evolution ends, we freeze $H^*$ and collect teacher trajectories, including reasoning and recovery branches. These are mixed with offline real-robot supervision to train tiny students (0.8B--4B), which replace the teacher in the same harness. Evaluation covers three training domains and three unseen benchmarks, plus three closed-loop real-robot tasks on an SO-ARM101 arm.}
\label{fig:overview}
\end{figure*}

TinyAstral has two stages (\cref{fig:overview}).
In the first, a master improves the harness of a frozen teacher through a statistically gated loop (\cref{sec:method:rsi}).
Once evolution ends, we freeze the final harness, collect teacher trajectories through it, and distill the resulting reasoning and recovery behavior into a compact student (\cref{sec:method:distill}). At deployment, the student replaces the teacher in the same harness.
We first define the agent and its interface (\cref{sec:method:setting}).

\subsection{Planner, harness, and scoring}
\label{sec:method:setting}

\paragraph{Agent and harness.}
A planner model $\pi$ controls a robot through a harness $H$.
At turn $t$ the harness gives $\pi$ an observation $o_t$, and $\pi$ produces a reasoning trace followed by exactly one tool call $a_t$. The harness executes the call and returns a structured result $r_t$.
The harness determines \emph{how} an action is carried out: perception (SAM~3 segmentation~\cite{carion2025sam} lifted to depth point clouds), grasp and placement geometry, collision-free motion planning, and closed-loop corrections.
The planner determines \emph{what} to do: which object, which part of it, which reference and spatial relation, which arm, in what order, and when the task is finished.
The model never outputs joint angles or coordinates unless a tool explicitly requests them.

Observations combine camera images, the task instruction, recent tool results, and gripper state. Object-level tools provide grasping, placement, verification, and recovery primitives; the full interface is specified in \cref{supp:harness_interface}.

\paragraph{Autonomous termination and scoring.}
The agent ends an episode itself by calling \texttt{done} with the predicates it believes hold (\eg \texttt{on(a,b)}, \texttt{upright(a)}, \texttt{lifted(a,cm)}). The harness checks these predicates with the cameras and gripper sensors; if a check fails, it rejects the call and returns the measurement so the agent can keep working. After two rejections, the next \texttt{done} ends the episode without a check.
An episode also ends if the agent calls \texttt{give\_up} or exhausts its budget (30 tool calls on RoboTwin; per-benchmark limits in \cref{supp:competitors}).
The environment's success flag never enters the model's input.
Our primary metric, strict success~$S$, counts an episode only if the agent ended it with \texttt{done} and the benchmark's success predicate holds in the final state. We also report \emph{ever} success, which counts an episode if the task was achieved after any tool call.

\subsection{Stage 1: Evolving the teacher harness}
\label{sec:method:rsi}

The self-improving system consists of the frozen teacher $\pi_T$, the harness $H$, and the master $M$.
Only $H$ changes. The weights of $\pi_T$ stay fixed throughout, and $M$ does not act in the environment.
Humans specify the rules (including the pre-registered acceptance criteria) and budget before the loop begins. The master is a set of frontier LLM agents that read trajectories, propose edits, implement them, and audit one another's code. The system applies the pre-registered paired statistical tests to determine acceptance.
Each round runs the current teacher, analyzes its failures, proposes edits, and tests a candidate against the current harness.

\paragraph{1. Read trajectories.}
The current accepted harness $H_k$ is run on development seeds. The master inspects the logged images, tool calls, tool results, and teacher reasoning, and assigns each failure to one of ten failure classes, such as off-target placement, false completion (\texttt{done} accepted but the final state fails), missing primitives, task-memory errors, and instance misidentification.

\paragraph{2. Propose edits.}
For a target cluster of failing tasks, the master writes candidate components: new tools or primitives, more honest tool returns, perception fixes, failure handling, or local corrections to the task memory.
Edits must be general.
Code may not branch on the task name. In the final integration phase, the system prompt and existing tool descriptions must remain byte-identical to the parent. Memory may change only for the targeted tasks.
In addition, every new component must fire through an explicit trigger that writes a component-specific marker to the log.
A separate code audit verifies that, whenever the trigger does not fire, the candidate behaves step-for-step like the parent.

\paragraph{3. Test and accept edits.}
Candidates are compared with the current harness on the same task/seed pairs. Training, selection, and evaluation seeds occupy disjoint ranges. The pre-registered gate requires positive evidence of improvement on targeted or triggered pairs and checks for regressions elsewhere; unsuccessful candidates are rejected or stopped early. The final integration phase accepted three of seven candidates.
Evaluation seeds are excluded from acceptance tests. The acceptance rules and capability ledger are in \cref{supp:rsi}.

\paragraph{Task memory.}
RoboTwin instructions often leave the success condition implicit (\eg ``lift'' requires holding the object at least 10\,cm up). For RoboTwin only, the harness therefore reads one short end-state note per task; the notes contain no actions or trajectories, are refined during RSI on development seeds, and are frozen and read-only at evaluation.
The other five benchmarks, including all zero-shot ones, use no task memory.

\paragraph{From evolution to a frozen teacher system.}
We number accepted main-line snapshots chronologically from the starting harness $H_0$; the lineage proceeds through $H_{16}$ to the final $H^*$ (\cref{tab:rsi_history}). These accepted versions are distinct from the 296 candidate capabilities and the seven integration cycles in the final phase. We freeze $H^*$ before collecting the final teacher trajectories for distillation.

\subsection{Stage 2: Collecting supervision from the frozen system}
\label{sec:method:distill}

The teacher $\pi_T$ collects trajectories through the frozen $H^*$. These trajectories supervise a student $\pi_S$ to reason about observations and feedback and choose the next tool call. The teacher weights and harness remain fixed during data collection and student training.
The distillation recipe combines reasoning targets, recovery branches, strict-success filtering, and training samples exported from the exact harness interactions.

\paragraph{Keep the teacher's reasoning.}
Each training target supervises the teacher's stated rationale together with its next tool call, \ie \texttt{<think>}$\ldots$\texttt{</think>} and then the call.
Emptying the reasoning text while keeping all samples identical lowers success by 19.3~pp on LIBERO-PRO, 11.2~pp on SimplerEnv Google VM, and 10.0~pp on VLABench (\cref{sec:exp:ablation}).

\paragraph{Recovery branches.}
Successful episodes rarely show how to recover from failure, so we add \emph{recovery branches}.
Starting from a successful teacher episode, we replay its first $k$ calls in a fresh simulator with the same seed, using the recorded results and without querying the model. At the branch point we inject a perturbation: on RoboTwin, the grasped object slips, or the target object is moved by 6--10\,cm before the next grasp; on Guava and LIBERO, a grasp misses, a push falls short, or a placed object is displaced.
The teacher then continues online.
A branch is kept only if the perturbation took effect, the replay reproduced every recorded result code, and the teacher still reached strict success. Its samples begin at the first decision after the perturbation.
On RoboTwin, branches are subsampled to 20\% of the RoboTwin samples (3{,}471 branch and 13{,}857 main-episode samples).

\paragraph{Strict successes only.}
Main episodes run at temperature $0$, as in evaluation. A training seed on which the teacher fails is retried up to twice at temperature $0.7$ with an unchanged prompt.
We keep only episodes that the teacher ended with \texttt{done} and that the benchmark judges successful in the final state. Episodes that succeeded at some point but were later undone are discarded.
Success labels are used offline for this selection and never appear in the training messages.

\paragraph{Offline real-robot supervision.}
To expose the student to real images, we add offline data from public real-robot datasets (AgiBot World~\cite{bu2025agibot}, RH20T~\cite{fang2024rh20t}; Fractal~\cite{brohan2023rt1}, BridgeData~V2~\cite{walke2023bridgedata}, DROID~\cite{khazatsky2024droid}, and Berkeley UR5~\cite{chen2023berkeleyur5} from Open X-Embodiment~\cite{oxe2024}), cast as three auxiliary tasks in the harness format: state and failure judgment (27{,}918 samples), sub-goal planning (4{,}187), and next tool call (9{,}771).
For next-call samples, we convert each recorded action sequence into a tool-call trajectory: gripper events and segment boundaries place the calls, and dataset annotations name the objects and spatial relations. Each step is rendered with the final harness's observation and history format, so the student sees real scenes through the same interface it uses in simulation.
The teacher then supplies the reasoning. It first predicts the call without seeing the reference; if the two match, its reasoning is kept. Otherwise, as in the rationalization step of STaR~\cite{zelikman2022star}, it is given the reference call and asked to explain it, and the explanation is kept only if it passes consistency and quality checks. In both cases the dataset-derived call is the action target (details in \cref{supp:real_supervision}).

\subsection{Student training and deployment}
\label{sec:method:training}
Teacher data comes from training seeds of all 50 RoboTwin tasks, the 7 seen Guava tasks, and LIBERO-90~\cite{liu2023libero}; the final set has 83{,}753 decision samples, half of them simulated (composition and hyper-parameters in \cref{supp:train}).
We fine-tune all language-model parameters of Qwen3.5-4B~\cite{qwen2026qwen35} for 2 epochs; at deployment the student replaces the teacher inside the frozen harness, served in bf16 on one accelerator.

\section{Experiments}
\label{sec:exp}

We first show how harness evolution improves frozen planners and how distillation transfers these gains to compact students. We then compare the final student with its teacher and other agents, and test its generalization to unseen benchmarks. Finally, we analyze training ingredients, recovery behavior, and inference compute cost.

\subsection{Setup}
\label{sec:exp:setup}

\paragraph{Benchmarks.}
We report six benchmarks.
Three belong to benchmark families that supply training data: RoboTwin~2.0~\cite{chen2025robotwin} with all 50 tasks $\times$ 10 evaluation seeds (500 episodes), Guava~\cite{liu2026guava} with 15 tasks $\times$ 30 official seeds (450 episodes; 7 tasks seen in training, 8 unseen), and LIBERO-PRO~\cite{zhou2025liberopro} with 300 episodes (three LIBERO~\cite{liu2023libero} suites, original and four perturbation types, 2 seeds each).
Three are never touched by RSI or training and test zero-shot transfer: SimplerEnv~\cite{li2024evaluating} (Visual Matching, 960 episodes on the Google Robot and WidowX), VLABench~\cite{zhang2024vlabench} (the first 500 episodes of the official list over five tracks), and EB-Manipulation~\cite{yang2025embodiedbench} (228 episodes).
RoboTwin is also the RSI development benchmark. Acceptance decisions use only selection seeds; evaluation seeds ($\geq 100{,}000$) are disjoint from selection and training seeds (\cref{sec:method:rsi}).
Guava evaluation seeds are disjoint from our Guava training seeds, and LIBERO-PRO training data comes only from LIBERO-90.

\paragraph{Evaluation protocol.}
We report strict success $S$ (\cref{sec:method:setting}) on the three training benchmarks and each benchmark's official success measure on the zero-shot benchmarks (\cref{tab:main}). VLABench stops at the official success event and EB-Manipulation latches success; neither defines a final-state predicate after \texttt{done}.
All rates use fixed denominators. Episodes with undetermined outcomes count as failures; upper bounds counting them as successes and \emph{ever} success are reported in \cref{tab:unified_full}.
Comparisons between systems we run are paired on identical (task, seed) episodes; published results are listed separately. We report success rates and paired differences in percentage points.
Within our harness, all evaluated planners decode at temperature~0 and share perception and motion-planning services. Competitor implementations and benchmark-specific configurations are detailed in \cref{supp:competitors}.

\subsection{Harness evolution and transfer to compact students}
\label{sec:exp:rsi}
\label{sec:exp:distill}
\begin{table}[t]
\centering
\footnotesize
\setlength{\tabcolsep}{3pt}
\begin{adjustbox}{max width=\linewidth}
\begin{tabular}{lrrrrr}
\toprule
Planner & $H_1$ & $H_7$ & $H_{16}$ & $H^*$ & Gain \\
\midrule
Qwen3.8-Flash-Next (teacher) & 33.4 & 49.4 & 60.4 & 63.6 & +30.2 \\
DeepSeek-V4.1-Flash & 32.8 & 45.0 & 57.6 & 63.4 & +30.6 \\
Qwen3.5-4B, untrained & 14.6 & 28.6 & 33.8 & 41.2 & +26.6 \\
\midrule
TinyAstral-4B & 36.2 & 49.8 & 60.0 & 62.8 & +26.6 \\
\bottomrule
\end{tabular}
\end{adjustbox}
\caption{\textbf{Transferring harness evolution gains across planners and into students.} Strict success (\%) on the same 500 RoboTwin episodes; gains are from $H_1$ to $H^*$ in percentage points. The first three rows keep model weights fixed across harness versions. The student row uses a separately trained model for each version (\cref{sec:exp:distill}).}
\label{tab:transfer}
\end{table}

\paragraph{Experimental design.}
We evaluate four harness versions, $H_1$, $H_7$, $H_{16}$, and $H^*$, on the same 500 RoboTwin episodes (\cref{tab:transfer}). The teacher, DeepSeek-V4.1-Flash, and the untrained 4B planner keep their weights fixed across versions. For each version, we collect teacher trajectories under the same collection-budget constraints and independently train a student from the same Qwen3.5-4B checkpoint on successful episodes, using the same training recipe (\cref{supp:train}). Each student runs in its corresponding harness.

\paragraph{Harness evolution.}
With teacher weights fixed, success rises from 33.4\% under $H_1$ to 63.6\% under $H^*$ (full lineage in \cref{tab:rsi_history}). The gains also transfer to other planners: DeepSeek-V4.1-Flash~\cite{deepseekai2026deepseekv41}, which was not the planner used during harness evolution, improves by 30.6~pp without model-specific harness changes. The untrained 4B planner improves from 14.6\% to 41.2\%. All three improve across the four measured versions.

\paragraph{Distilling evolution gains.}
We train a separate 4B student for each harness version. Student success increases from 36.2\% at $H_1$ to 62.8\% at $H^*$, with each student scoring close to its corresponding teacher (\cref{tab:transfer}). Thus, improvements to the teacher's harness carry into the distilled students. Within the final harness $H^*$, the distilled student also outperforms the untrained 4B model by 21.6 percentage points (62.8\% vs.\ 41.2\%), showing the additional benefit of distillation.

\subsection{Performance and generalization}
\label{sec:exp:main}
\label{sec:exp:zeroshot}
\begin{table*}[t]
\centering
\footnotesize
\setlength{\tabcolsep}{4pt}\def\cY{\checkmark}\def\cN{\setlength{\fboxsep}{0.6pt}\colorbox{black!18}{\makebox[0.9em]{$\times$}}}\def\cU{--}
\begin{adjustbox}{max width=\linewidth}
\begin{tabular}{@{}l cccc c ccc|cccc@{}}
\toprule
 & \multicolumn{5}{c}{Evaluation conditions} & \multicolumn{3}{c|}{Training benchmarks ($S$ unless noted)} & \multicolumn{4}{c}{Zero-shot benchmarks (official measure)} \\
\cmidrule(lr){2-6}\cmidrule(lr){7-9}\cmidrule(l){10-13}
Method & Open & \makecell{No closed\\LLM} & \makecell{No succ.\\signal} & \makecell{No ICL\\demo} & \makecell{No expert\\demo} & \makecell{RoboTwin-50\\500 ep.} & \makecell{Guava\\450 ep.} & \makecell{LIBERO-PRO\\300 ep.} & \makecell{SimplerEnv\\Google VM} & \makecell{SimplerEnv\\WidowX} & \makecell{VLABench\\500 ep.} & \makecell{EB-Manip.\\228 ep.} \\
\midrule
\multicolumn{13}{@{}l}{\emph{Published results (their protocols and episodes)}} \\
GPT-6 Astra, one-shot~\cite{guo2026transferring} & \cN & \cN & \cN & \cN & \cY & 73.6 & -- & -- & -- & -- & -- & -- \\
PyRUA-Lean, GPT-6 Astra~\cite{si2026fewer} & \cN & \cN & \cN & \cY & \cN & 69.2 & -- & -- & -- & -- & -- & -- \\
GPT-6 Astra, zero-shot~\cite{guo2026transferring} & \cN & \cN & \cN & \cY & \cY & 53.2 & -- & -- & -- & -- & -- & -- \\
GPT-5.4, Guava harness~\cite{liu2026guava} & \cN & \cN & \cY & \cY & \cY & -- & 90.4 & -- & -- & -- & -- & -- \\
MotorMind, GPT-6 Sol~\cite{li2026motormind} & \cN & \cN & \cN & \cY & \cY & -- & -- & 83.3\rlap{$^{a}$} & -- & -- & -- & -- \\
World Action Agent, Gemini 3.7 Flash~\cite{zhang2026world} & \cN & \cN & \cN & \cY & \cN & -- & -- & 75.6\rlap{$^{e}$} & -- & -- & -- & -- \\
World Action Agent, Qwen3.5-9B student~\cite{zhang2026world} & \cN & \cN & \cN & \cY & \cN & -- & -- & 43.3\rlap{$^{e}$} & -- & -- & -- & -- \\
SoFar, GPT-4o~\cite{qi2025sofar} & \cN & \cN & \cY & \cY & \cY & -- & -- & -- & 74.9 & 58.3 & -- & -- \\
$\pi_0$-FAST, fine-tuned~\cite{pertsch2025fast} & \cY & \cY & \cY & \cY & \cN & -- & -- & -- & 61.9 & 32.1 & 43.1 & -- \\
GPT-4o, EmbodiedBench$^{b}$~\cite{yang2025embodiedbench} & \cN & \cN & \cY & \cN & \cY & -- & -- & -- & -- & -- & -- & 28.9 \\
\midrule
\multicolumn{13}{@{}l}{\emph{Same episodes, competitors' official harnesses, run by us}} \\
\multicolumn{13}{@{}l}{\hspace{0.5em}\textit{\scriptsize Open 4B agents}} \\
PyRUA-Lean-4B~\cite{si2026fewer} & \cY & \cY & \cN & \cY & \cN & 46.0$^{f}$ & -- & 60.0$^{f}$ & -- & -- & -- & -- \\
Guava-4B~\cite{liu2026guava} & \cY & \cY & \cY & \cY & \cY & -- & 84.4 & -- & -- & -- & -- & -- \\
\multicolumn{13}{@{}l}{\hspace{0.5em}\textit{\scriptsize Same model as our teacher (Qwen3.8-Flash-Next), competitors' harnesses}} \\
MotorMind + Qwen3.8-Flash-Next$^{c}$~\cite{li2026motormind} & \cY & \cY & \cN & \cY & \cY & -- & -- & 57.3 & -- & -- & -- & -- \\
Guava harness + Qwen3.8-Flash-Next~\cite{liu2026guava} & \cY & \cY & \cY & \cY & \cY & -- & 78.9 & -- & -- & -- & -- & -- \\
\multicolumn{13}{@{}l}{\hspace{0.5em}\textit{\scriptsize Open harness-RSI model}} \\
SenseNova-RoboRSI-27B$^{d}$~\cite{deng2026sensenova} & \cY & \cY & \cN & \cY & \cY & -- & -- & 70.3 & -- & -- & -- & -- \\
\midrule
\multicolumn{13}{@{}l}{\emph{Ours (same harness $H^*$, strict protocol)}} \\
Qwen3.5-4B, untrained & \cY & \cY & \cY & \cY & \cY & 41.2\rlap{$^{\dagger}$} & 75.1 & 36.3\rlap{$^{\dagger}$} & 56.6 & 86.5 & 27.6 & 21.9 \\
Teacher: Qwen3.8-Flash-Next (180B) & \cY & \cY & \cY & \cY & \cY & 63.6 & 94.7 & 80.0 & 69.2 & 92.7 & 46.2 & 32.0 \\
DeepSeek-V4.1-Flash (552B) & \cY & \cY & \cY & \cY & \cY & 63.4 & -- & 70.3\rlap{$^{\dagger}$} & 68.0 & 86.5 & 38.8 & 29.8 \\
TinyAstral-0.8B & \cY & \cY & \cY & \cY & \cY & 61.8 & 93.8 & 54.7 & 64.0 & 87.5 & 40.4 & 24.6 \\
TinyAstral-2B & \cY & \cY & \cY & \cY & \cY & 59.8 & 88.0 & 69.0 & 65.4 & 88.5 & 42.0 & 25.0 \\
\textbf{TinyAstral-4B} & \cY & \cY & \cY & \cY & \cY & \textbf{62.8} & \textbf{93.6} & \textbf{78.3} & \textbf{66.6} & \textbf{88.5} & \textbf{41.2} & \textbf{27.6} \\
\bottomrule
\end{tabular}
\end{adjustbox}
\caption{\textbf{Performance on six benchmarks} (success \%). Published results and same-episode runs are grouped separately. We use strict success on training benchmarks and official metrics on zero-shot benchmarks (\cref{sec:exp:setup}).
\checkmark/$\times$: condition satisfied/not satisfied; --: unavailable. ICL: in-context demonstrations; expert demo: expert demonstrations from the evaluated benchmark. No succ.\ signal: no benchmark success signal provided to the planner.}
\label{tab:main}
\vspace{2pt}
\begin{minipage}{\linewidth}
\scriptsize\raggedright
\textit{Notes.} $^{a}$Original LIBERO suites, single seed. $^{b}$Ground-truth coordinates, boxes, and in-context examples. $^{c,d}$Stop on success. $^{e}$Position and task perturbations only (360 episodes), simulator point clouds; the student is trained on rollouts of the same tasks and tested on unseen seeds.
$^{f}$Official code and settings (per-step success flag, 90-minute budget) on our episodes.
$^{\dagger}$${>}5\%$ unknown outcomes, counted as failures; bounds in \cref{tab:unified_full}. Full protocols: \cref{supp:competitors,supp:published}.
\end{minipage}
\end{table*}

\begin{figure*}[tp]%
\centering
\includegraphics[width=\textwidth]{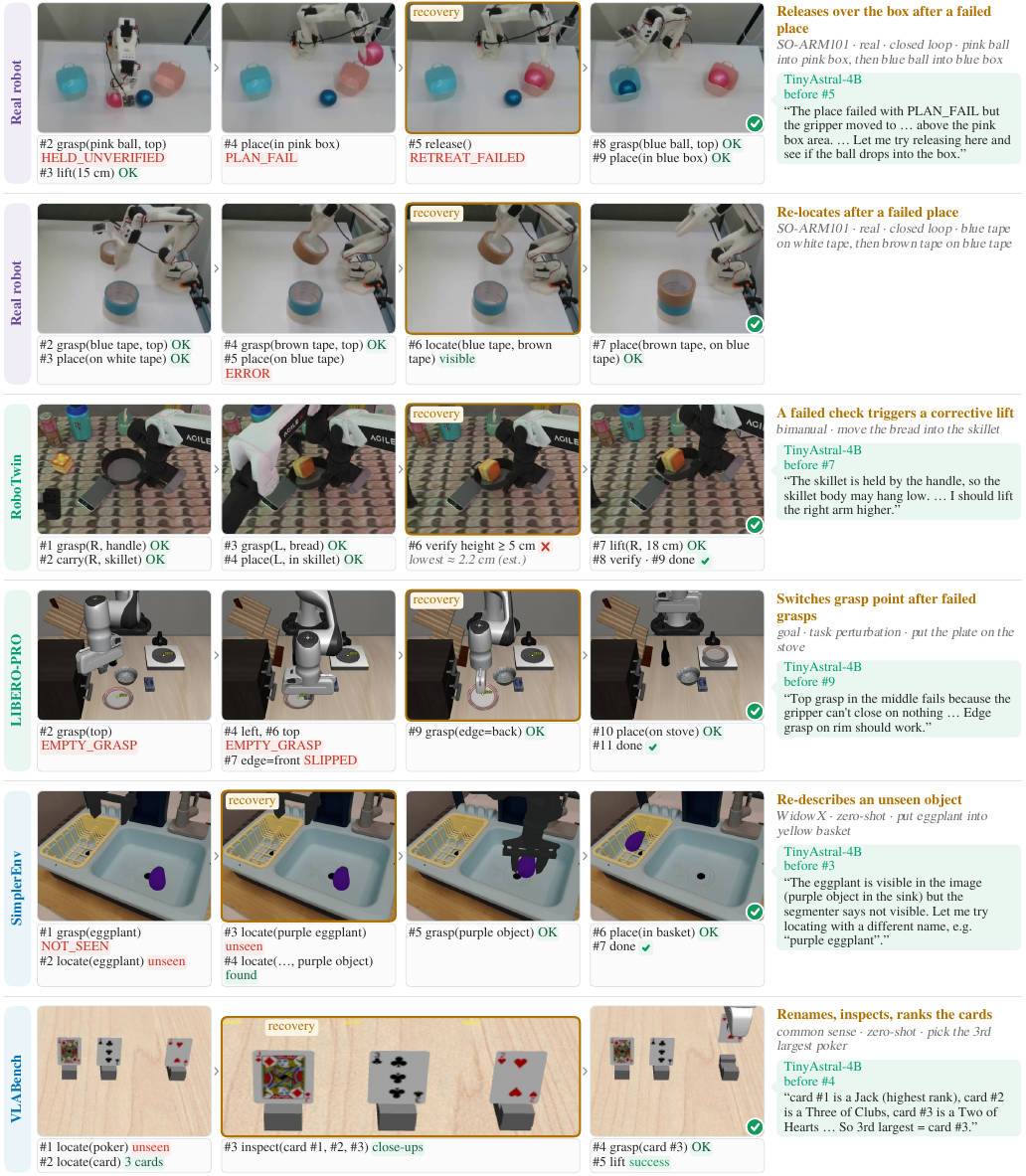}
\caption{\textbf{\oursfour recovers on its own} (original evaluation runs, final harness $H^*$; blue labels: zero-shot). Top two rows: closed loop on a real SO-ARM101 arm; middle two: training domains; bottom two: zero-shot. Cards: tool calls and returned codes verbatim from the logs; amber: the decisive step; bubbles: the student's reasoning before it. More examples, including Guava and a distilled-vs-untrained comparison on one episode, are in \cref{fig:qualitative_suppA,fig:qualitative_suppB}.}
\label{fig:qualitative}
\end{figure*}

\paragraph{Student versus teacher.}
TinyAstral-4B stays close to its teacher on all three training benchmarks: 62.8\% vs.\ 63.6\% on RoboTwin, 93.6\% vs.\ 94.7\% on Guava, and 78.3\% vs.\ 80.0\% on LIBERO-PRO (\cref{tab:main}). In the same harness, it also improves over the untrained 4B model by 18.4~pp on Guava and 42.0~pp on LIBERO-PRO; counting all undetermined outcomes as successes raises the untrained model's LIBERO-PRO score to at most 57.3\%.

\paragraph{Comparison with other agents.}
On the same evaluation episodes, TinyAstral-4B outperforms the tested open agents, including Guava-4B, PyRUA-Lean-4B, and SenseNova-RoboRSI-27B (\cref{tab:main}). It also exceeds systems using our teacher model, Qwen3.8-Flash-Next, in other harnesses: 93.6\% vs.\ 78.9\% for the Guava harness, and 78.3\% vs.\ 57.3\% for MotorMind on LIBERO-PRO. PyRUA-Lean-4B, run in its official setting with a per-step success signal that our protocol withholds, reaches 46.0\% on RoboTwin and 60.0\% on LIBERO-PRO. These results highlight the contribution of the evolved harness and distilled policy to the final system's performance. Competitor-specific execution and scoring protocols are detailed in \cref{supp:competitors}.

\paragraph{Generalization to unseen benchmarks.}
TinyAstral-4B also generalizes to three benchmarks never used for RSI or training: SimplerEnv, VLABench, and EB-Manipulation. Its success rates are within 5.0~pp of its teacher and within 2.4~pp of the 552B DeepSeek-V4.1-Flash, all using the same harness (\cref{tab:main}). Compared with the untrained 4B model, distillation improves performance across all four evaluation settings, with gains of 10.0~pp on SimplerEnv Google VM and 13.6~pp on VLABench. Despite receiving no training on these benchmarks, the student also exceeds published results from $\pi_0$-FAST~\cite{pertsch2025fast} on both SimplerEnv robots and $\pi_0$~\cite{black2024pi0} on VLABench, although these baselines were fine-tuned on in-domain data (protocol details in \cref{supp:published}).

\paragraph{Smaller students.}
TinyAstral-0.8B reaches 61.8\% on RoboTwin and 93.8\% on Guava, close to the 4B student, but falls to 54.7\% on LIBERO-PRO compared with 78.3\%. On the zero-shot benchmarks, both the 0.8B and 2B students remain within 3.0~pp of the 4B model across all four settings. The effect of model size is therefore most pronounced on LIBERO-PRO among the evaluated benchmarks (\cref{tab:main}).

\subsection{Ablations and recovery behavior}
\label{sec:exp:ablation}
\begin{table}[t]
\centering
\footnotesize
\setlength{\tabcolsep}{4pt}
\begin{adjustbox}{max width=\linewidth}
\begin{tabular}{lccc}
\toprule
 & Full recipe & $-$ teacher thinking & $-$ real-robot data \\
\midrule
\multicolumn{4}{l}{\emph{Training benchmarks (strict $S$)}} \\
\quad RoboTwin-50 & \textbf{62.8} & 62.4 {\scriptsize($-$0.4)} & 63.4 {\scriptsize(+0.6)} \\
\quad Guava & \textbf{93.6} & 88.9 {\scriptsize($-$4.7)} & 95.1 {\scriptsize(+1.6)} \\
\quad\quad unseen tasks & \textbf{91.7} & 82.9 {\scriptsize($-$8.8)} & 91.7 \\
\quad LIBERO-PRO & \textbf{78.3} & 59.0 {\scriptsize($-$19.3)} & 78.3 {\scriptsize(0.0)} \\
\midrule
\multicolumn{4}{l}{\emph{Zero-shot benchmarks (official measure)}} \\
\quad SimplerEnv Google VM & \textbf{66.6} & 55.5 {\scriptsize($-$11.2)} & 66.3 {\scriptsize($-$0.4)} \\
\quad SimplerEnv WidowX & \textbf{88.5} & 87.5 {\scriptsize($-$1.0)} & 91.7 {\scriptsize(+3.1)} \\
\quad VLABench & \textbf{41.2} & 31.2 {\scriptsize($-$10.0)} & 45.2 {\scriptsize(+4.0)} \\
\quad EB-Manipulation & \textbf{27.6} & 23.2 {\scriptsize($-$4.4)} & 28.9 {\scriptsize(+1.3)} \\
\midrule
Real-robot next call (offline) & \textbf{0.923} & -- & 0.235 {\scriptsize($-$0.688)} \\
\bottomrule
\end{tabular}
\end{adjustbox}
\caption{\textbf{Distillation recipe ablations on the final 4B student} (success \%; in parentheses: paired difference to the full recipe). Each variant changes one ingredient only.
}
\label{tab:ablation}
\end{table}

\paragraph{Teacher reasoning.}
Keeping the teacher's reasoning has the largest measured effect among our ablations (\cref{tab:ablation}). Removing the reasoning text while keeping all training samples otherwise unchanged lowers success by 19.3~pp on LIBERO-PRO, 11.2~pp on SimplerEnv Google VM, and 10.0~pp on VLABench. On LIBERO-PRO, 49 of 300 episodes end at the call limit without reasoning, compared with 7 under the full recipe.

\paragraph{Auxiliary real-robot supervision.}
Adding real-robot data raises next-call accuracy on held-out real-robot episodes from 0.235 to 0.923, with no statistically detectable change in success across the seven simulated settings (\cref{tab:ablation}). Its measured benefit is in offline decisions on real observations.

\paragraph{Behavior under failure.}
The student responds to failures by re-describing target objects, correcting placements, and changing grasp points, in both training and unseen benchmarks (\cref{fig:qualitative,fig:qualitative_suppA,fig:qualitative_suppB}). The same student also runs in closed loop on a real SO-ARM101 arm, sorting two balls into matching boxes and stacking three tape rolls (\cref{fig:qualitative}, top). Of the 271 zero-shot episodes solved by the student but not by the untrained 4B model in the same harness, 171 (63\%) end with the untrained model exhausting its call budget. Additional examples, including a failure, are shown in \cref{fig:qualitative_suppA,fig:qualitative_suppB}.

\subsection{Inference compute cost}
\label{sec:exp:cost}
\begin{table}[t]
\centering
\footnotesize
\setlength{\tabcolsep}{2.5pt}
\begin{adjustbox}{max width=\linewidth}
\begin{tabular}{lcccccc}
\toprule
 & & \multicolumn{2}{c}{Median s / episode} & \multicolumn{2}{c}{Accel.$\cdot$s / episode} & \\
\cmidrule(lr){3-4}\cmidrule(lr){5-6}
Model & Accel. & no spec. & spec. & deployed & vs.\ 4B & Succ. \\
\midrule
\textbf{TinyAstral-4B} (bf16) & 1 & 25.3 & -- & \textbf{25.3} & \textbf{1.0$\times$} & 14/20 \\
Qwen3.8-Flash-Next teacher (FP8, TP4) & 4 & 55.0 & 31.0 & 123.8 & 4.9$\times$ & 12/20 \\
DeepSeek-V4.1-Flash (TP8) & 8 & 38.1 & 31.2 & 249.6 & 9.9$\times$ & 13/20 \\
\bottomrule
\end{tabular}
\end{adjustbox}
\caption{\textbf{Deployment cost on 20 shared SimplerEnv episodes.} Time is median wall-clock duration, including simulation and perception; Accel.$\cdot$s equals planner accelerator count $\times$ time. All planners run on the same type of accelerator. Deployed costs use speculative decoding (``spec.'') for the large models and none for the student. ``Succ.'' reports successes without speculative decoding.}
\label{tab:cost}
\end{table}

We measure inference compute cost as planner accelerators multiplied by median episode duration, using the same 20 SimplerEnv episodes for each system.
In their deployed configurations, the student uses 25.3\,accelerator$\cdot$s against 123.8 for the teacher (4.9$\times$) and 249.6 for DeepSeek-V4.1-Flash (9.9$\times$), and its median episode is also the shortest (25.3\,s vs.\ 31.0 and 31.2\,s; \cref{tab:cost}).
The 4B student is served in bf16 on one accelerator without quantization or speculative decoding; the teacher uses four accelerators with FP8 weights and tensor parallelism, and DeepSeek-V4.1-Flash uses eight.

\section{Conclusion}
\label{sec:conclusion}
TinyAstral first improves a frozen teacher through harness evolution, then distills its reasoning and recovery behavior into a compact student deployed in the final harness. Across four harness versions, students trained from later versions perform better and stay close to their corresponding teachers, showing that harness improvements carry into compact agents. The harness gains also transfer to other frozen planners.
The final 4B student outperforms every open competitor we ran on identical episodes, stays close to its teacher on three training benchmarks, scores close to a 552B planner on unseen benchmarks, and uses about one fifth of the teacher's accelerator time in our deployment measurement. These results support harness evolution followed by distillation as a route to capable robot agents with lower inference compute cost.
Future work will extend the approach to more RSI environments.

{
    \small
    \bibliographystyle{ieeenat_fullname}
    \bibliography{refs}      %
}

\ifdefined\NOSUPP\else
\clearpage
\setcounter{page}{1}
\maketitlesupplementary
\appendix

\section{All runs}
\label{supp:unified}
\begin{table}[t]
\centering
\scriptsize
\setlength{\tabcolsep}{3pt}
\begin{adjustbox}{max width=\linewidth}
\begin{tabular}{llrrrrr}
\toprule
Benchmark & Run & $N$ & $S/N$ & $(S{+}U)/N$ & $U$ & ever \\
\midrule
RoboTwin-50 & Teacher, $H_1$ & 500 & 33.4 & 34.6 & 6 & 41.0 \\
RoboTwin-50 & Teacher, $H^*$ & 500 & 63.6 & 65.4 & 9 & 68.8 \\
RoboTwin-50 & DeepSeek-V4.1-Flash & 500 & 63.4 & 66.8 & 17 & 68.6 \\
RoboTwin-50 & Qwen3.5-4B, untrained & 500 & 41.2 & 47.6 & 32 & 51.4 \\
RoboTwin-50 & TinyAstral-0.8B & 500 & 61.8 & 63.0 & 6 & 68.2 \\
RoboTwin-50 & TinyAstral-2B & 500 & 59.8 & 62.8 & 15 & 67.4 \\
RoboTwin-50 & TinyAstral-4B & 500 & 62.8 & 65.2 & 12 & 68.0 \\
RoboTwin-50 & TinyAstral-4B, $H_1$ student & 500 & 36.2 & 38.4 & 11 & 42.2 \\
\midrule
Guava & Teacher, $H^*$ & 450 & 94.7 & 94.7 & 0 & 95.3 \\
Guava & Qwen3.5-4B, untrained & 450 & 75.1 & 75.1 & 0 & 89.6 \\
Guava & TinyAstral-0.8B & 450 & 93.8 & 93.8 & 0 & 94.2 \\
Guava & TinyAstral-2B & 450 & 88.0 & 88.0 & 0 & 90.9 \\
Guava & TinyAstral-4B & 450 & 93.6 & 93.6 & 0 & 94.0 \\
Guava & Guava-4B & 450 & 84.4 & 84.4 & 0 & 86.7 \\
\midrule
LIBERO-PRO & Teacher, $H^*$ & 300 & 80.0 & 85.0 & 15 & 83.0 \\
LIBERO-PRO & DeepSeek-V4.1-Flash & 300 & 70.3 & 80.7 & 31 & 72.3 \\
LIBERO-PRO & Qwen3.5-4B, untrained & 300 & 36.3 & 57.3 & 63 & 41.7 \\
LIBERO-PRO & TinyAstral-0.8B & 300 & 54.7 & 57.7 & 9 & 60.3 \\
LIBERO-PRO & TinyAstral-2B & 300 & 69.0 & 73.7 & 14 & 70.7 \\
LIBERO-PRO & TinyAstral-4B & 300 & 78.3 & 80.3 & 6 & 81.3 \\
\bottomrule
\end{tabular}
\end{adjustbox}
\caption{\textbf{All runs on the training benchmarks} (\%); definitions in \cref{sec:method:setting,sec:exp:setup}.}
\label{tab:unified_full}
\end{table}

\Cref{tab:unified_full} lists all runs on the training benchmarks; definitions are in \cref{sec:method:setting,sec:exp:setup}.

\section{Competitor protocols}
\label{supp:competitors}
\paragraph{Our budgets.} Every planner run in our harness uses the same per-episode tool-call limit on a given benchmark (\cref{tab:budgets}). Guava uses its official per-task round limits; the other limits were set by us before evaluation.
\begin{table}[h]
\centering
\footnotesize
\begin{tabular}{lcl}
\toprule
Benchmark & Tool calls & Basis \\
\midrule
RoboTwin & 30 & set by us \\
Guava & 25 / 30 & official per-task limits~\cite{liu2026guava} \\
LIBERO-PRO & 40 & set by us \\
SimplerEnv & 25 & set by us, as Guava \\
VLABench & 40 & set by us, as LIBERO-PRO \\
EB-Manipulation & 25 & set by us, as Guava \\
\bottomrule
\end{tabular}
\caption{Per-episode tool-call limits of our harness.}
\label{tab:budgets}
\end{table}

\paragraph{Competitors.} All competitors run from their released code and weights on our accelerators; no closed-source API is used.
\textbf{Guava-4B}~\cite{liu2026guava} runs in the official Guava harness (weights at revision \texttt{19a8c77}, official prompt and parser, temperature 0) on our 450 Guava episodes, with the same per-task round limits; its result is re-scored with our unified metric (380/450).
\textbf{PyRUA-Lean-4B}~\cite{si2026fewer} runs with its official code and settings (per-step success flag, 90-minute budget) on our 500 RoboTwin and 300 LIBERO-PRO episodes (230/500 and 180/300).

\paragraph{Same teacher in alternative systems.}
Guava harness with Qwen3.8-Flash-Next uses the official prompt and parser with medium reasoning effort and default sampling, whereas our teacher uses its default xhigh reasoning and temperature~0. Both allow 25/30 calls per task. MotorMind with this planner uses its official stop-on-success protocol and uncapped calls (median 77, compared with 8 for our student on LIBERO-PRO).

\paragraph{Additional competitor metrics.}
SenseNova-RoboRSI-27B runs with its official stop-on-success rule, 40 policy calls, and an empty task-memory package, and reaches 211/300.

\section{Published comparisons}
\label{supp:published}
\begin{table*}[t]
\centering
\scriptsize
\setlength{\tabcolsep}{2pt}\def\cY{\checkmark}\def\cN{\setlength{\fboxsep}{0.6pt}\colorbox{black!18}{\makebox[0.9em]{$\times$}}}\def\cU{--}
\begin{tabular}{@{}p{0.28\linewidth} ccccc p{0.35\linewidth} c@{}}
\toprule
Published system & Open & \makecell{No closed\\LLM} & \makecell{No succ.\\signal} & \makecell{No ICL\\demo} & \makecell{Train\\data} & Advantages over our setting & Result \\
\midrule
\multicolumn{8}{@{}l}{\emph{RoboTwin~2.0, 50 tasks (ours: 62.8)}} \\
GPT-6 Astra, one-shot (RoboDawn~\cite{guo2026transferring}) & \cN & \cN & \cN & \cN & --- & closed model; one demo per task; stops on success & 73.6 \\
PyRUA-Lean, GPT-6 Astra~\cite{si2026fewer} & \cN & \cN & \cN & \cY & E & closed model; frozen RoboTwin-SFT VLA executor; success flag; guides & 69.2 \\
Harness VLA, GPT-5.5~\cite{zhang2026harness}$^{l}$ & \cN & \cN & \cN & \cN & E & closed model; RoboTwin-tuned VLA; ICL trace; ever-success & 62.4 \\
Gemini-3.8-Flash, one-shot (RoboDawn~\cite{guo2026transferring}) & \cN & \cN & \cN & \cN & --- & closed model; one demo per task; stops on success & 62.2 \\
GPT-6 Astra, tool calling (PyRUA-Lean~\cite{si2026fewer}) & \cN & \cN & \cN & \cY & E & closed model; frozen RoboTwin-SFT VLA executor; success flag & 60.0 \\
Harness VLA, Claude Code~\cite{zhang2026harness} & \cN & \cN & \cN & \cN & E & as Harness VLA, GPT-5.5 & 58.4 \\
GPT-6 Astra, zero-shot (RoboDawn~\cite{guo2026transferring}) & \cN & \cN & \cN & \cY & --- & closed model; stops on success; checker verdict on \texttt{done} & 53.2 \\
PyRUA-Lean Coder-4B, self-reported~\cite{si2026fewer} & \cY & \cY & \cN & \cY & E & success flag; frozen RoboTwin-SFT VLA executor (coder trained on LIBERO-PRO only) & 52.4 \\
LingBot-VLA / $\pi_{0.5}$ (RoboDawn~\cite{guo2026transferring}) & \cY & \cY & \cN & \cY & E & 50 demos per task; stops on success & 50.4 / 46.0 \\
Robo-Harness K1 + Gemini~\cite{li2026robo} & \cN & \cN & \cN & \cY & --- & closed model; 10-task subset & 32.0 / 28.0 \\
\midrule
\multicolumn{8}{@{}l}{\emph{Guava, 15 tasks (ours: 93.6)}} \\
GPT-5.4 in the Guava harness~\cite{liu2026guava} & \cN & \cN & \cY & \cY & --- & closed model & 90.4 \\
Guava-4B, self-reported~\cite{liu2026guava} & \cY & \cY & \cY & \cY & T & own protocol & 87.1 \\
\midrule
\multicolumn{8}{@{}l}{\emph{LIBERO-PRO (ours: 78.3)}} \\
MotorMind, GPT-6 Sol~\cite{li2026motormind} & \cN & \cN & \cN & \cY & --- & closed model; original suites; stops on success & 83.3 \\
World Action Agent, Gemini 3.7 Flash~\cite{zhang2026world} & \cN & \cN & \cN & \cY & E & closed model; stops on success; skills from LIBERO-90 expert demos; simulator point clouds; position/task splits only & 75.6 \\
MotorMind, Qwen3.8-Flash-Next~\cite{li2026motormind} & \cY & \cY & \cN & \cY & --- & stops on success; 20 episodes per suite & 66.7 / 53.8 \\
ASPIRE~\cite{lu2026aspire}$^{l}$ & \cN & \cN & \cN & \cY & --- & closed model; simulator-verified search & 61.4 \\
GPT-5.4 / Guava-4B in the Guava harness~\cite{liu2026guava} & \cN\,/\,\cY & \cN\,/\,\cY & \cU & \cY & ---\,/\,T & CaP-X protocol; GPT-5.4 closed & 45.0/44.0; 41.0/41.0 \\
World Action Agent, Qwen3.5-9B student~\cite{zhang2026world} & \cN & \cN & \cN & \cY & T+E & Gemini sub-agents and grounding; trained on teacher rollouts of the same tasks (unseen seeds); stops on success & 43.3 \\
CaP-X, 10 loops (as run in~\cite{li2026motormind})~\cite{fu2026capx} & \cN & \cN & \cU & \cY & --- & closed models; 10 refinement loops & 13.3 / 19.2 \\
\midrule
\multicolumn{8}{@{}l}{\emph{SimplerEnv, visual matching (ours: Google VM 66.6, WidowX 88.5)}} \\
SoFar (GPT-4o)~\cite{qi2025sofar} & \cN & \cN & \cY & \cY & O & closed model; orientation module & 74.9 / 58.3 \\
$\pi_0$ / $\pi_0$-FAST, Google VM~\cite{black2024pi0,pertsch2025fast}$^{v}$ & \cY & \cY & \cY & \cY & E & fine-tuned on Fractal & 58.8 / 61.9 \\
$\pi_0$ / $\pi_0$-FAST, WidowX~\cite{black2024pi0,pertsch2025fast}$^{v}$ & \cY & \cY & \cY & \cY & E & fine-tuned on Bridge & 27.1 / 32.1 \\
Best fine-tuned VLA, WidowX~\cite{starvla2026} & \cY & \cY & \cY & \cY & E & fine-tuned on Bridge & 65.3 \\
\midrule
\multicolumn{8}{@{}l}{\emph{VLABench (ours: 41.2)}} \\
$\pi_0$ / $\pi_0$-FAST~\cite{black2024pi0,pertsch2025fast} & \cY & \cY & \cY & \cY & E & fine-tuned on VLABench~\cite{vlabench2025pi0card,vlabench2025pi0fastcard} & 29.4 / 43.1 \\
SHAPER~\cite{wang2026shaper} & \cY & \cY & \cY & \cY & E & VLABench-tuned executor; harness evolved on target & 34.5 \\
\midrule
\multicolumn{8}{@{}l}{\emph{EB-Manipulation (ours: 27.6)}} \\
GPT-4o / Claude-3.7-Sonnet~\cite{yang2025embodiedbench} & \cN & \cN & \cY & \cN & --- & closed models; GT coordinates and boxes; ICL examples & 28.9 / 28.5 \\
22 further models in EmbodiedBench~\cite{yang2025embodiedbench} & \cU & \cU & \cY & \cN & --- & GT coordinates and boxes; ICL examples & 26.8 \\
Robot-R1~\cite{kim2025robotr1} & \cN & \cY & \cY & \cN & O & privileged EB inputs & 11.7 \\
\bottomrule
\end{tabular}
\caption{\textbf{Published results on our six benchmarks} (success \%), each reported by the cited work under its own protocol and episodes; the TinyAstral-4B value of \cref{tab:main} is given in each block header.
Condition columns as in \cref{tab:main} (\colorbox{black!18}{$\times$}: advantage used by the system); Train data: --- none, T teacher rollouts, E expert demonstrations, O other datasets.
$^{l}$Official RPent leaderboard (\url{https://rpent.readthedocs.io/en/latest/rst_source/leaderboard/performance.html}, accessed 2026-10-23). $^{v}$As collected in StarVLA~\cite{starvla2026}.
}
\label{tab:published}
\end{table*}

\Cref{tab:published} lists published results on our six benchmarks together with the advantages each system uses over our setting.

\section{More qualitative examples}
\label{supp:qualitative}
\begin{figure*}[t]
\centering
\includegraphics[width=\textwidth]{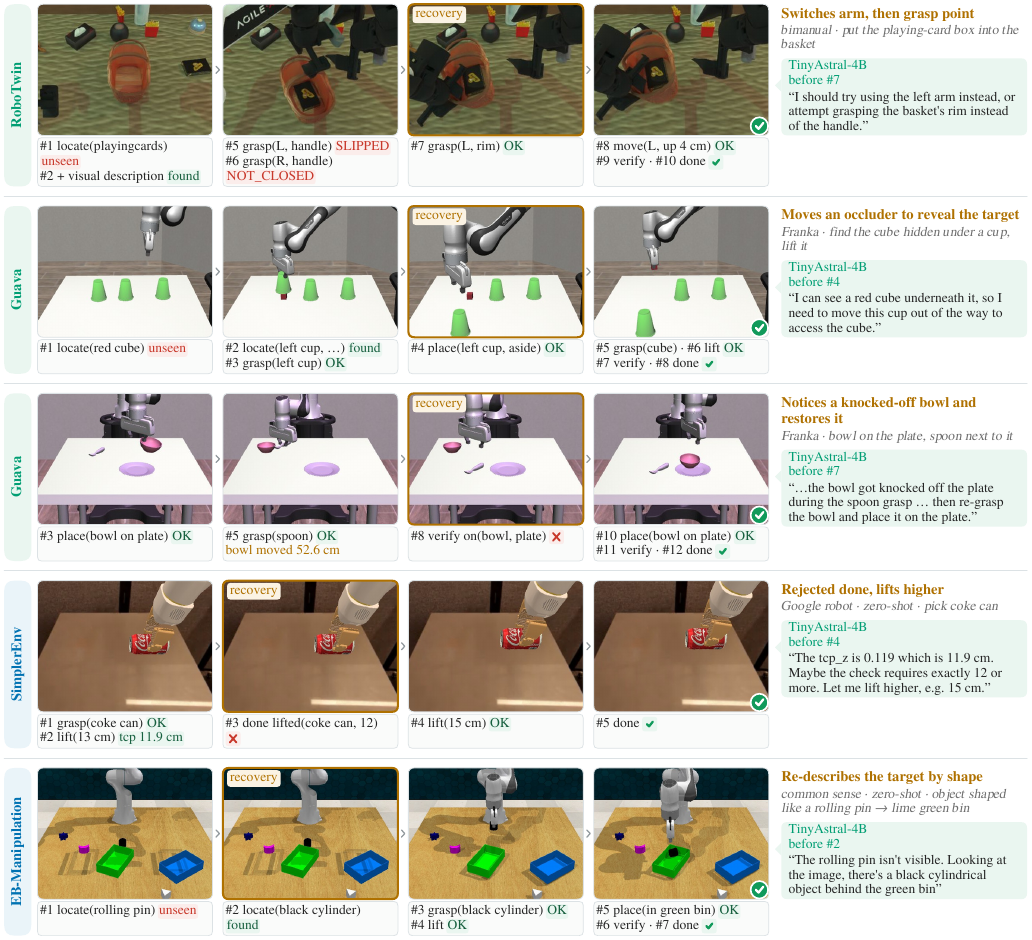}
\caption{\textbf{More recoveries of \oursfour} (conventions as in \cref{fig:qualitative}).
(1)~RoboTwin: the card box is found only by a visual description; the basket handle slips with either arm, so the student grasps the rim.
(2)~Guava: the student lifts a cup to reveal the hidden cube, sets the cup aside and picks up the cube.
(3)~Guava: notices a knocked-off bowl and restores it.
(4)~SimplerEnv Google robot (zero-shot): \texttt{done} is rejected at 11.9\,cm, below the student's own 12\,cm check; it lifts to 15\,cm and finishes.
(5)~EB-Manipulation (zero-shot): ``rolling pin'' is not found, so the student describes it by shape (``black cylinder'').}
\label{fig:qualitative_suppA}
\end{figure*}

\begin{figure*}[t]
\centering
\includegraphics[width=\textwidth]{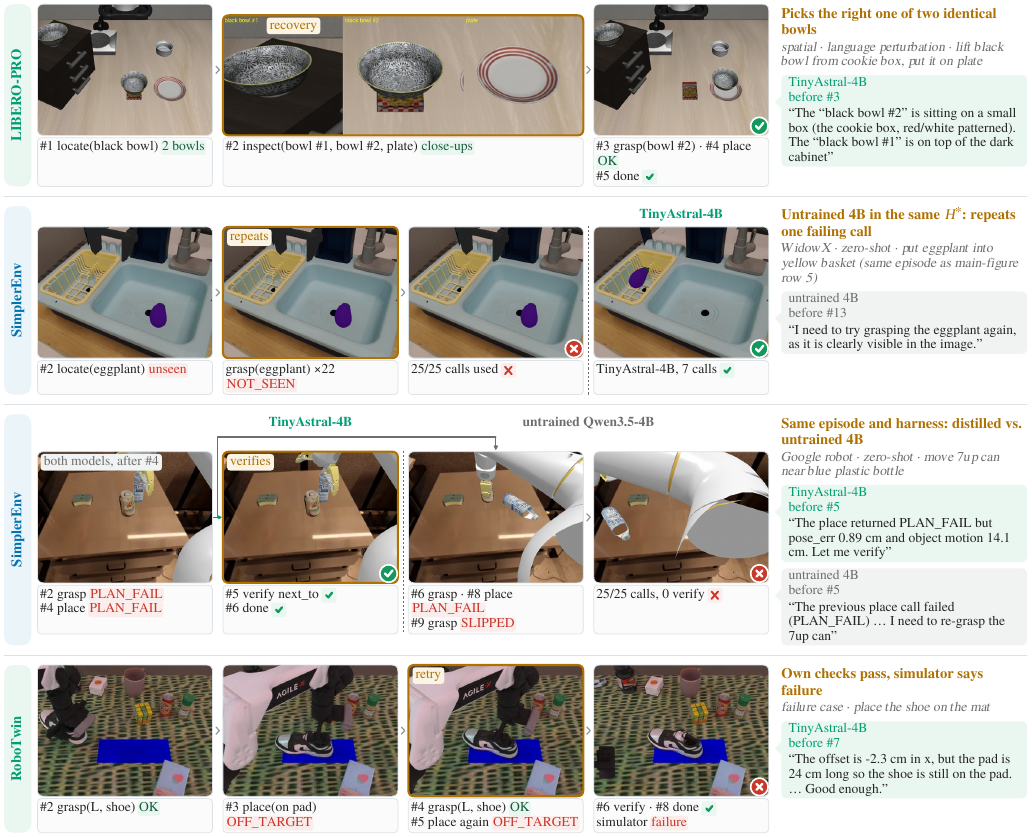}
\caption{\textbf{Choosing the right object, the untrained baseline, and a failure.}
(1)~LIBERO-PRO, language perturbation: of two identical black bowls, the student picks the one on the cookie box after inspecting close-ups.
(2)~The untrained Qwen3.5-4B in $H^*$ on the episode of \cref{fig:qualitative}, row~5, repeats \texttt{grasp(eggplant)} 22 times (\texttt{NOT\_SEEN}) until the budget runs out; the student re-describes the object and succeeds in 7 calls.
(3)~SimplerEnv Google robot (zero-shot), same episode and harness: distilled vs.\ untrained 4B.
(4)~A student failure on RoboTwin: the shoe ends \texttt{OFF\_TARGET} by 2.3\,cm, the student's own checks pass and it declares success, but the simulator judges a failure.}
\label{fig:qualitative_suppB}
\end{figure*}

\Cref{fig:qualitative_suppA,fig:qualitative_suppB} extend \cref{fig:qualitative}.

\section{RSI details}
\label{supp:rsi}
\begin{table}[t]
\centering
\footnotesize
\setlength{\tabcolsep}{4pt}
\begin{tabular}{lrrrrrrr}
\toprule
Harness & $H_1$ & $H_2$ & $H_7$ & $H_{12}$ & $H_{15}$ & $H_{16}$ & $H^*$ \\
\midrule
Success & 33.4 & 33.6 & 49.4 & 50.8 & 54.6 & 60.4 & 63.6 \\
\bottomrule
\end{tabular}
\caption{\textbf{Accepted harness snapshots.} Teacher success (\%) on the 500 RoboTwin-50 evaluation episodes (fixed denominator) for accepted harness snapshots; $H_k$ is the $k$-th accepted snapshot and $H^*$ the frozen final version.}
\label{tab:rsi_history}
\end{table}

\Cref{tab:rsi_history} lists the accepted snapshots of the main lineage. Their acceptance rules and the capability ledger follow.

\paragraph{Direction screen.}
Candidates first pass a cheap \emph{direction screen} on small seed blocks. A candidate is discarded if the task-averaged 95\% upper bound of its gain on the target cluster falls below $+2$~pp.
Surviving components are merged into one candidate $H'$. $H'$ and its parent $H_k$ then run concurrently, interleaved on the same $\sim$1{,}150 pre-registered (task, seed) pairs, so that the two arms share load and infrastructure conditions.

\paragraph{Statistical verdict.}
Let $J$ denote the pairs in the target cluster together with the non-target pairs in which a candidate marker fired, and let $U$ denote the remaining pairs.
$H'$ is accepted if (i) the one-sided 97.5\% lower bound of the paired gain on $J$, computed by a task-stratified bootstrap (seeds resampled within each task, tasks weighted equally), is positive, and (ii) the net number of wins on the target cluster is positive.
$H'$ is vetoed if the untriggered set $U$ regresses by at least 2~pp with a two-sided 95\% upper bound below zero.
At the halfway point, the test stops for futility if the net gain on $J$ is non-positive or the conditional power falls below 10\%.
The number of pairs is fixed in advance.

\paragraph{Capability ledger.}
Every candidate is recorded in a ledger with the failure that triggered it, its proposer, its acceptance evidence, and its fate.
Over the whole run, the master proposed 296 candidate capabilities. Of these, 121 were adopted, 16 were adopted and later superseded, 142 were rejected, and 17 remained undecided. The final harness contains 101 capabilities, 87 verbatim and 14 in a successor form; almost all adopted capabilities missing from it were accepted on side branches that were never merged into the main line.

\section{Training details}
\label{supp:train}
The 41{,}877 simulated samples comprise 13{,}857 RoboTwin, 9{,}054 Guava, and 11{,}159 LIBERO main-episode samples, plus 3{,}471, 2{,}048, and 2{,}288 recovery-branch samples, respectively.
Each sample is one decision, with the loss applied only to the final assistant turn.
The vision encoder and projector stay frozen. Training uses learning rate $10^{-5}$ with cosine decay and 5\% warm-up, a global batch of 32, and a maximum length of 12{,}288 tokens. The 2B and 0.8B students are trained on the same data with the same recipe.

\paragraph{Real-robot next-call pool.}
\label{supp:real_supervision}
Explanations given the reference call are rejected if they refer to the supplied answer, claim unmeasured success, omit the target object or required reference, or violate the length and format constraints.
Next-call relabeling retained 19{,}598 of 23{,}070 candidate steps: 4{,}509 whose reasoning comes from a matching first prediction and 15{,}089 explained given the reference call. The 9{,}771 next-call training samples are drawn from this pool.

\section{Harness interface details}
\label{supp:harness_interface}
\paragraph{Observations.}
On RoboTwin, $o_t$ contains RGB images from the head camera and both wrist cameras, the task instruction together with the names of scene objects, a proprioceptive line with the measured gripper state of each arm, the result $r_{t-1}$ of the previous call, and a counter of calls used.
Images are kept for the two most recent turns only. Earlier turns retain their text, and earlier assistant messages retain only the tool call, without reasoning.
On RoboTwin the first turn also includes the task note (\cref{sec:method:rsi}).
Guava and LIBERO-PRO use the same planner loop and tool semantics through benchmark adapters, keeping the three most recent single-view images.

\paragraph{Tools.}
The RoboTwin harness exposes object-level tools: \texttt{locate}, \texttt{grasp} (with an optional part such as handle, rim, or end), \texttt{place} and \texttt{carry} (relative to a reference object, with optional facing or turning), \texttt{handover}, \texttt{lift}, \texttt{articulate}, \texttt{press}, \texttt{push}, \texttt{pour}, \texttt{move}, \texttt{release}, \texttt{home}, \texttt{bimanual} (two calls executed in parallel), \texttt{verify}, \texttt{done}, and \texttt{give\_up}.
Tools report what they measured rather than what they intended. For example, \texttt{place} re-measures the object after release and returns \texttt{OFF\_TARGET} with the offset if the object missed its target. A check whose value cannot be measured reliably is returned as unknown (\texttt{ok=null}) rather than as a pass.

\fi

\end{document}